\def\year{2019}\relax
\pdfoutput=1
\documentclass[letterpaper]{article} 
\usepackage{aaai19}  
\usepackage{times}  
\usepackage{helvet}  
\usepackage{courier}  
\usepackage{url}  
\usepackage{graphicx}  
\usepackage{subfig}
\usepackage{multirow}
\usepackage{amsfonts,amssymb}
\begin{document}
%
\title{M2M-GAN: Many-to-Many Generative Adversarial Transfer Learning for Person Re-Identification}
\author{Wenqi Liang, Guangcong Wang, Jianhuang Lai, Junyong Zhu\\
School of Data and Computer Science, Sun Yat-sen University, China\\
}
\maketitle
\begin{abstract}
Cross-domain transfer learning (CDTL) is an extremely challenging task for the person re-identification (ReID). Given a source domain with annotations and a target domain without annotations, CDTL seeks an effective method to transfer the knowledge from the source domain to the target domain. However, such a simple two-domain transfer learning method is unavailable for the person ReID in that the source/target domain consists of several sub-domains, e.g., camera-based sub-domains. To address this intractable problem, we propose a novel Many-to-Many Generative Adversarial Transfer Learning method (M2M-GAN) that takes multiple source sub-domains and multiple target sub-domains into consideration and performs each sub-domain transferring mapping from the source domain to the target domain in a unified optimization process. The proposed method first translates the image styles of source sub-domains into that of target sub-domains, and then performs the supervised learning by using the transferred images and the corresponding annotations in source domain. As the gap is reduced, M2M-GAN achieves a promising result for the cross-domain person ReID. Experimental results on three benchmark datasets Market-1501, DukeMTMC-reID and MSMT17 show the effectiveness of our M2M-GAN.
\end{abstract}
\section{Introduction}
Person re-identification aims to retrieve person images across non-overlapping camera views given a probe image. Recently, a wide range of deep models, which uses classification loss \cite{Deng_2018_CVPR,zhuo2018occluded}, triplet-based loss \cite{Wang2016DARI,ding2015deep,wang2017tcsvt} and verification loss \cite{li2014deepreid,chen2015deep}, achieve a significant improvement for the person ReID community. To further improve the performance, a lot of methods even exploit extra expensive annotations, e.g., keypoint annotations \cite{Su2017ICCV,Zhao_2017_CVPR}, attribute annotations and semantic segmentation annotations \cite{Kalayeh_2018_CVPR,Song_2018_CVPR}. However, when applied to a new specific scenario, these methods require annotators to re-label a large amount of annotations to obtain a good performance. Motivated by this key point, an intuitive question arises: can we exploit knowledge from an existing labeled dataset (labeled source domain) and then transfer it to a target scenario (unlabeled target dataset)?

\begin{figure}[!t]
\centering
\includegraphics[width=2.8in]{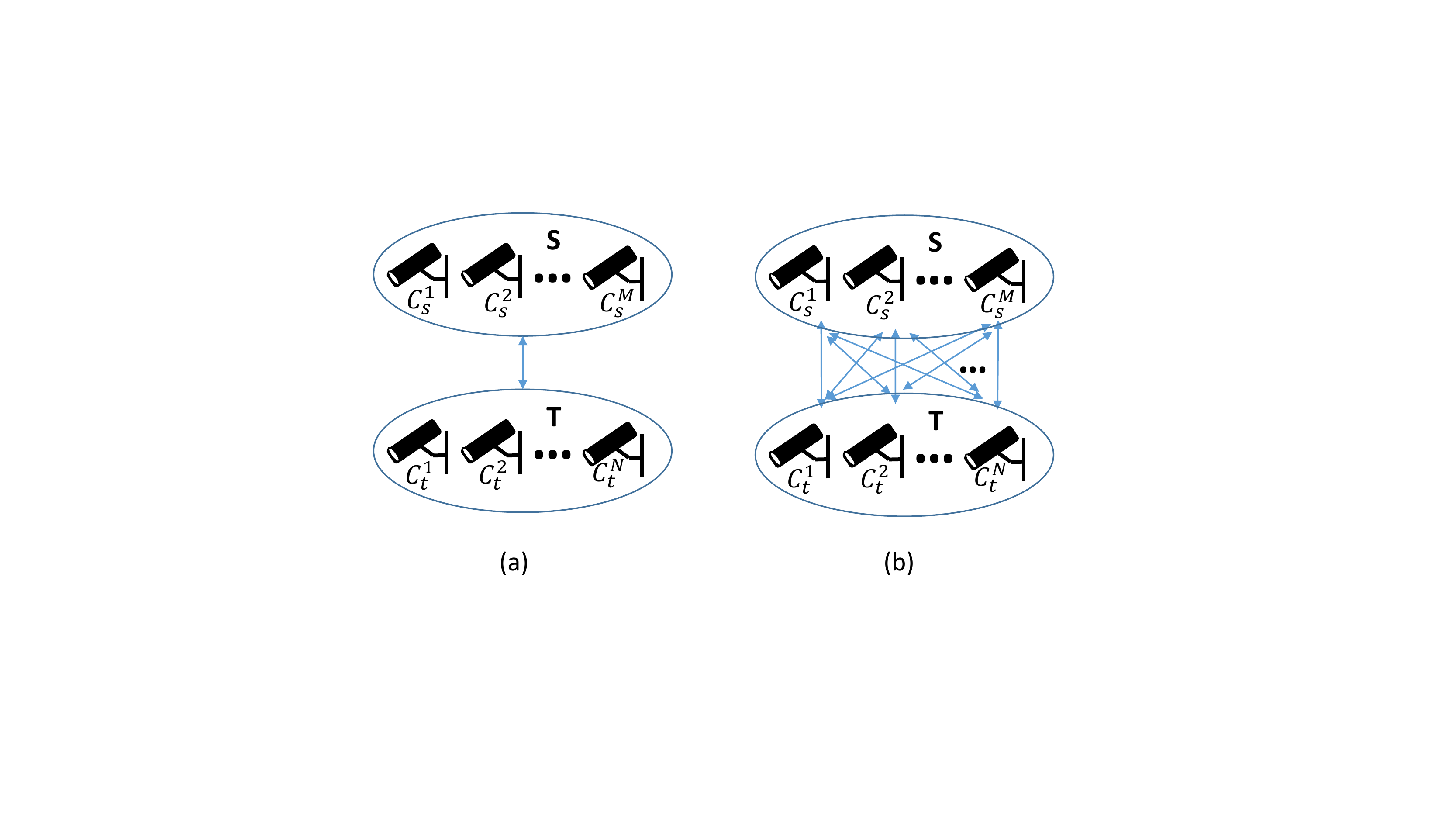}
\caption{Classical one-to-one cross-domain transfer learning method vs. Our M2M-GAN method. A source domain $S$ contains $M$ camera-based sub-domains and a target domain $T$ contains $N$ ones. (a) Coarse one-to-one domain transferring method: it directly performs the transfer learning from a source domain to a target domain without considering camera-based sub-domains. (b) Our many-to-many transferring method (M2M-GAN): all pairs of sub-domain transferring relationships are optimized in a unified process by sharing parameters.}
\label{fig:define}
\end{figure}

A common approach for this question is to pre-train a model in a source domain and directly apply it to a target domain. However, there is a large gap between the source domain and the target domain in person ReID due to different lighting conditions, backgrounds, viewpoints, etc. Can we narrow the domain gap by exploiting the relationship between the source domain and the target domain? Recent cross-domain person ReID approaches address this issue by using a two-stage approach, i.e., transferring the labeled source domain to the target domain by using generative adversarial networks (GANs) and then performing supervised learning by using labeled transferred images. So, how to translate the labeled source domain to the target domain (the first stage) is the key point to the cross-domain transfer learning (CDTL) of person ReID.

Existing CDTL methods of person ReID \cite{Deng_2018_CVPR,Wei_2018_CVPR} simply formulates CDTL as a classical one-to-one cycle generative adversarial model (CycleGAN) \cite{CycleGAN2017}, which is employed for image style transferring between two domains, as shown in Figure \ref{fig:define} (a). However, one-to-one CDTL models neglect the fact that the source/target domain naturally consists of many sub-domains, i.e., camera-based sub-domains. In a source/target domain (dataset), feature representations across disjoint camera views of the same person follow different distributions due to the changes of viewpoints, lighting conditions and camera features \cite{chen2015mirror}. The ignorance of different camera-based sub-domains confuses GANs (GANs do not know which sub-domain should be generated) and thus deteriorates the domain adaptation performance.

Inspired by this point, one would think distinguishing different camera-based sub-domains in the source/target domain may be beneficial for a CDTL ReID system. Given a CDTL ReID system that contains $M$ source sub-domains and $N$ target sub-domains, a natural approach to this problem is training $M\times N$ separate CycleGANs, respectively. As an initial attempt, Wei et al. \cite{Wei_2018_CVPR} tried to transfer a large-scale dataset to a small-scale dataset with two cameras. To simplify, they do not distinguish different source sub-domains and only consider two target sub-domains by using two CycleGANs for CDTL. When applied to a large-scale dataset (e.g., MSMT17 with 15 cameras), they pointed out the expensive computational cost of separate CycleGANs for all pairs of source-to-target sub-domains and directly use classical one-to-one CDTL to reduce the model's complexity on large-scale datasets.

Considering the drawbacks of these cross-domain ReID methods, we propose a novel many-to-many generative adversarial transfer learning method (M2M-GAN) that takes multiple source sub-domains and multiple target sub-domains into consideration and performs each sub-domain transferring mapping from the source domain to the target domain in a unified optimization process, as shown in Figure \ref{fig:define} (b). To accomplish this, the source sub-domain label $l_s^i$ and the target sub-domain label $l_t^j$ are embedded into an image $x_{s_i}$, which guides a generator to translate the image $x_{s_i}$ from the source sub-domain $S_i$ to the target sub-domain $T_j$ and vice versa. With the guided information,  M2M-GAN can easily address the many-to-many CDTL for person ReID.


Compared with $M\times N$ separate CycleGANs, our M2M-GAN method includes several fundamental properties. First, $M\times N$ pair-wise sub-domain transferring mappings (GANs) are naturally integrated into one M2M-GAN model with $1/(M \times N)$ parameters. For example, the Market-1501 dataset contains 6 cameras and the MSMT17 dataset contains 15 cameras, and thus there are 90 transferring mappings in all. M2M-GAN only needs $1/90$ of the parameters compared with $M\times N$ separate CycleGANs. Second, the training time of M2M-GAN is also significantly reduced due to the joint optimization. Third, M2M-GAN provides a good solution for many-to-many CDTL in person ReID. M2M-GAN jointly optimizes $M\times N$ transferring mappings that share the similar transferring knowledge while separate CycleGANs neglect this key point and have limited data (e.g., the dataset is split into $M\times N$ pairs of source-to-target sub-sets) for the learning of each transferring mapping, which may encounter the over-fitting problem.

Overall, this paper makes three main contributions:
\begin{itemize}
  \item First, as far as we know, it is the first attempt to address the many-to-many cross-domain transfer learning (CDTL) problem for person ReID. We highlight that many-to-many CDTL is much more appropriate for cross-domain person ReID than one-to-one CDTL because feature representations across disjoint camera views of the same person follow different distributions.

  \item Second, to solve the many-to-many CDTL problem for person ReID, we propose a novel many-to-many generative adversarial transfer learning (M2M-GAN), which translates the image style from source sub-domains to target sub-domains with less training time, fewer parameters and a better performance. Moreover, we also integrate a mask-based identity preserve loss, a camera-based sub-domain classification loss into the proposed framework.

  \item Third, extensive experiments show that the proposed method is effective in three benchmark datasets, i.e., Market-1501, DukeMTMC-reID and MSMT17.
\end{itemize}

\section{Related Work}
Recent deep models have shown remarkable success in person ReID by using different types of loss, e.g., classification loss, verification loss and triplet loss. For example, generalized similarity measure methods \cite{Wang2016DARI,lin2016cross} are naturally integrated into deep networks. Feng et al. \cite{feng2018learning} proposed a view-specific person ReID framework by leveraging the classification loss, center loss and Euclidean distance constraint. Ding et al. \cite{ding2015deep} proposed an image-based triplet loss to reduce the computational cost. Based on the image-based triplet loss, Wang et al. \cite{wang2017tcsvt} extended a kNN-triplet for the image-to-video person ReID.

Recently, a rich variety of person ReID methods pay attention to extra information to further improve the performance. For example, Li et al. \cite{Li_2017_CVPR} used the global body-based feature, latent person parts and local part-based features for person representation. Zhao et al. \cite{Zhao_2017_ICCV} proposed to decompose the human body into regions (parts) for person matching and aggregate the similarities computed between the corresponding regions as the overall matching score. Su et al. \cite{Su2017ICCV} introduced a pose-driven deep convolution model to alleviate the pose variations and learn robust feature representations from both the global images and different local parts. Zhao et al. \cite{Zhao_2017_CVPR} introduced a SpindleNet based on human body region guided multi-stage feature decomposition and tree-structured competitive feature fusion. Song et al. \cite{Song_2018_CVPR} introduced the binary segmentation masks to construct synthetic RGB-Mask pairs as inputs, and designed a mask-guided contrastive attention model
to learn features separately from the body and background regions. Kalayeh \cite{Kalayeh_2018_CVPR} et al. attempted to integrate human semantic parsing into person ReID and achieved a competitive result. Wang et al. \cite{wang2018st_ReID} proposed a st-ReID method with a spatial-temporal constraint, which dramatically outperformed current state-of-the-arts.

Instead of focusing on supervised person ReID, some studies attempt to transfer the knowledge from a labeled domain to an unlabeled domain, which arises in many real-world person ReID applications. Cross-domain transfer learning (CDTL) methods in person ReID can be categorized into two groups, i.e., single-stage transferring and two-stage transferring. Single-stage CDTL methods directly build a relationship between a source domain and a target domain at the feature level. For example, Ganin et al. \cite{Ganin2016jrml} performed the domain adaption by adding a domain classifier and a gradient reversal layer.
\begin{figure*}[!t]
\centering
\includegraphics[width=5.5in]{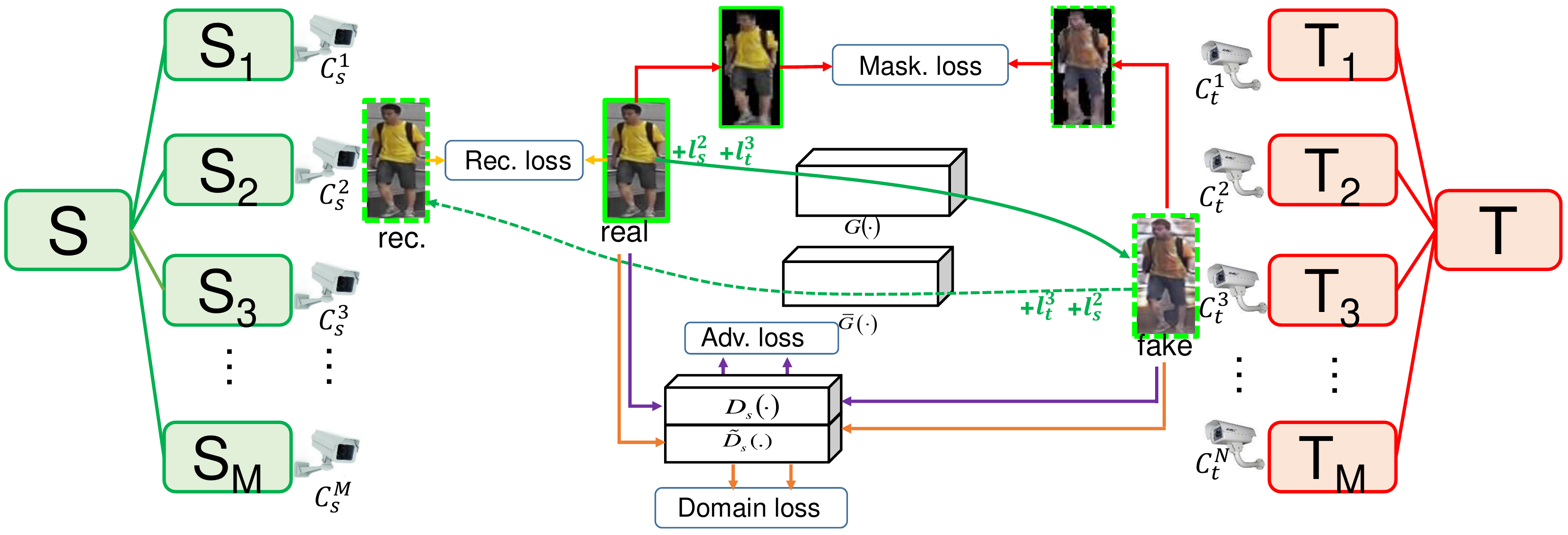}
\caption{The proposed M2M-GAN architecture. For simplicity, it only describes a pair of sub-domain transferring from a source sub-domain $S_i$ to a target sub-domain $T_j$. Given a source image $x_{s_i}$ with a source sub-domain label $l_{s_i}$ and a target sub-domain label $l_{t_i}$ that is the desired transferred image style, we use cycle loss, adversarial loss, identity loss and reconstruction loss to train M2M-GAN and obtain a good transferred (fake) image $x_{t_i}^{*}$ in the target sub-domain. The same process are performed during the target-to-source image style transferring. (Best viewed in color)}
\label{fig:overview}
\end{figure*}
Differently, two-stage CDTL methods first generate new training images by translating the data distribution of the source domain to that of the target domain at the image level. Then the transferred images with source domain annotations are used for supervised learning. For example, Deng et al. \cite{Deng_2018_CVPR} proposed to preserve self-similarity of an image before and after translation and domain-dissimilarity of a translated source image and a target image. However, this method simply regards the person ReID transfer learning problem as a one-to-one domain adaption problem but neglects the fact that a source/target domain contains several sub-domains. Wei et al. \cite{Wei_2018_CVPR} proposed a person transfer method to bridge domain gap in a one-to-one manner and also implicitly used two separate CycleGANs for a small-scale dataset transferring, which can be regarded as a one-to-two CDTL. However, they did not distinguish different source sub-domains and pointed out that training many separate CycleGANs is quite expensive and unavailable for large-scale datasets that contains lots of cameras. Our approach differs from these methods, as the proposed model explicitly formulate the cross-domain person ReID problem as a many-to-many adversarial transfer learning problem.

Concurrently, other unsupervised domain adaptation methods \cite{Zhong_2018_CVPR,zhong2018generalizing}, simply using CycleGAN \cite{CycleGAN2017} or StarGAN \cite{Choi_2018_CVPR}, are also developed. These methods focus on the image style translation between target sub-domains while the images in the source domain are ignored or simply used to form negative pairs for model training. This pipeline pays attention to the intra-domain generative adversarial transfer learning while our M2M-GAN method aims for the inter-domain generative adversarial transfer learning. 

\section{Proposed Method}
\label{sec:met}
\subsection{Problem Definition.}
Cross-domain transfer learning (CDTL) of the person ReID is featured in two aspects. First, it contains a labeled source domain and an unlabeled target domain. The goal of CDTL in person ReID is to transfer the distribution of the source domain to that of the target domain conceptually. Second, the source/target domain contains several sub-domains, i.e., camera-based sub-domains. Different camera-based sub-domains in the same domain have something in common but differ a lot from each other due to different lighting conditions, backgrounds, and viewpoints.

Formally, let $S$ denote a labeled source domain and $T$ denote an unlabeled target domain. $S$ contains $M$ sub-domains, denoting $S_1,S_2,...,S_i,...,S_M$. The corresponding cameras are $C_s^1,C_s^2,...,C_s^i,...,C_s^M$. $T$ contains $N$ sub-domains, denoting $T_1,T_2,...,T_j,T_N$. The corresponding cameras are $C_t^1,C_t^2,...,C_t^j,...,C_t^N$. The goal of many-to-many CDTL in person ReID is to translate the image style from $S_i$ to $T_j$ given any $i$ and $j$, where $1 \le i \le M$ and $1 \le j \le N$. Let a labeled image $(x_{s_{i}},y_{s_{i}}) \in S_i$. The subscript of $x$ denotes which sub-domain the image follows. Many-to-many CDTL translates the real image $x_{s_i}$ to the fake image $x_{t_j}^{*}$ such that $x_{t_j}^{*}$ follows the distribution of $T_j$. The fake image $x_{t_j}^{*}$ and the corresponding label $y_{s_{i}}$ are used to supervised learning for person ReID.

\subsection{Formulation of M2M-GAN.}
Supervised person ReID methods \cite{chen2015mirror,feng2018learning} have revealed that feature representations across disjoint camera views of the same person follow different distributions due to the changes of viewpoints, lighting conditions and camera features. Directly using CycleGAN to transfer the image styles between $S$ and $T$ will confuse the generators because $S$ and $T$ contain different sub-domains and generators do not know which source sub-domain or target sub-domain should be generated. Without considering the existing of sub-domains, CycleGAN can hardly offer an optimal solution for the many-to-many CDTL problem.

To solve this problem, a M2M-GAN is proposed in this paper. The main idea behind our M2M-GAN is to embed the source sub-domain label information $l_{s_i}$ and the target sub-domain label information $l_{t_j}$ into the image $x_{s_i}$ and guide the generators to generate the desired image $x_{t_j}^*$. By doing this, generators know which target sub-domain it will generate and which source sub-domain it comes from during the source-to-target training process (note that the target-to-source training process is the same). With the guided feature maps, M2M-GAN can jointly optimize $M\times N$ transferring mappings from the source sub-domains to the target sub-domains in a unified process. For simplicity, we only consider the source-to-target translation. We embed this guided information into an image by
\begin{equation}
\label{eq:embed}
{x_{embed}^{s_it_j}} = [x_{rgb},B_s^i,B_t^j]
\end{equation}
where $[ . , . ]$ represents channel concatenation, $x_{rgb}$ represents a RGB image of size $(3,h,w)$, $B_s^i$ represents a binary tensor of size $(M,h,w)$, $B_t^j$ represents a binary tensor of size $(N,h,w)$, and $x_{embed}^{s_it_j}$ denotes a tensor of size $(M+N+3,h,w)$. The $i$-th channel of $B_s^i$ is ``one" value, denoting that $x_{rgb}$ comes from the $i$-th source sub-domain and the remaining $M-1$ channels are ``zero" values. The $j$-th channel of $B_t^j$ is ``one" value, denoting that $x_{rgb}$ will be translated to the $j$-th target sub-domain and the remaining $N-1$ channels are ``zero" values. In this way, $x_{embed}^{s_it_j}$ guides M2M-GAN to translate $x_{rgb}$ into a 3-channel fake image $x_{rgb}^*$ from $S_i$ to $T_j$. In the similar way, we obtain $x_{embed}^{t_js_i}$.

Overall, we use two generators, two discriminators and two domain classifiers, among which half of them are used for the source-to-target transferring while the rest of them are used for the target-to-source transferring. We use a cycle loss, an adversarial loss, an identity loss and a reconstruction loss to constrain the networks to obtain a good transferred (fake) image $x_{t_i}^{*}$ for supervised learning. Figure \ref{fig:overview} shows the overview of our M2M-GAN for the many-to-many cross-domain person ReID. Details.

\textbf{\emph{Adversarial Loss.}}
To translate the distribution of the source sub-domain $S_i$ to that of the target sub-domain $T_j$, the adversarial loss is used
\begin{equation}
\label{eq:adv}
\begin{array}{*{20}{l}}
   {{L_{adv}}(G,{D_t})  } & {=\sum\limits_{i = 1}^M {\sum\limits_{j = 1}^N {}}({\mathbb{E}_{{x_{t_j}}}}[\log {D_t}({x_{t_j}})]  }  \\
   {} & {+{\mathbb{E}_{{x_{s_i}}}}[\log (1 - {D_t}(G({x_{s_i}},l_s^i,l_t^j)))])}  \\
\end{array}
\end{equation}
where $G$ attempts to generate a fake image $G({x_{s_i}},l_s^i,l_t^j)$ that is indistinguishable from the images in $T_j$ given an image $x_{s_i}$ and domain labels $l_s^i$ and $l_t^j$ while $D_t$ attempts to distinguish between real and fake images in $T_j$. Mathematically, $G$ attempts to minimize ${{L_{adv}}(G,{D_t})}$ while $D_t$ attempts to maximize ${{L_{adv}}(G,{D_t})}$.


In the same way, we obtain a similar adversarial loss for the transferring mapping from the target sub-domain $T_j$ to the source sub-domain $S_i$, i.e., ${L_{adv}}({\bar G},{D_s})$, where ${\bar G}$ denotes a generator which is used for the target-to-source transferring and $D_s$ is used to distinguish between real and fake images in $S_i$.

\textbf{\emph{Reconstruction Loss.}} Reconstruction loss, also implicitly termed cycle loss, is widely used in unsupervised auto-encoder. It aims to regularize the mappings such that ${x_{s_i}} \to G({x_{s_i}},l_s^i,l_t^j) \to \bar G(G({x_{s_i}},l_s^i,l_t^j),l_t^j,l_s^i) \approx {x_{s_i}}$ and ${x_{t_j}} \to \bar G({x_{t_j}},l_t^j,l_s^i) \to G(\bar G({x_{t_j}},l_t^j,l_s^i),l_s^i,l_t^j) \approx {x_{t_j}}$. Considering these two constraints, we compute the reconstruction loss by
\begin{equation}
\label{eq:rec}
\begin{array}{*{20}{l}}
   {{L_{rec}(G,\bar G)}} & {=\sum\limits_{i = 1}^M {\sum\limits_{j = 1}^N {}}(||{x_s} - \bar G(G({x_s},l_s^i,l_t^j),l_t^j,l_s^i)|{|_1} }  \\
   {} & {+||{x_t} - G(\bar G({x_t},l_t^j,l_s^i),l_s^i,l_t^j)|{|_1)}}  \\
\end{array}
\end{equation}

\textbf{\emph{Mask-based Identity Preserve Loss.}} To further preserve the content of the person images, we use Mask-RCNN to perform person segmentation and attempt to preserve the foreground of the images before and after generating fake images. Without the identity preserve loss, it is hard for M2M-GAN to preserve the content of person images due to large cross-domain variations, e.g., different person identities, significant changes in illumination, background clutter, pose, viewpoint, and occlusion. We compute the mask-based identity preserve loss by
\begin{equation}
\label{eq:mask}
\begin{array}{*{20}{l}}
   {} & {{L_{mask}(G,\bar G)}}  \\
    =  & {\sum\limits_{i = 1}^M {\sum\limits_{j = 1}^N {}}(||{x_{{s_i}}} \bullet M({x_{{s_i}}}) - G({x_{{s_i}}},l_s^i,l_t^j) \bullet M({x_{{s_i}}})|{|_2} + }  \\
   {} & {||{x_{{t_j}}} \bullet M({x_{{t_j}}}) - \bar G({x_{{t_j}}},l_t^j,l_s^i) \bullet M({x_{{t_j}}})|{|_2)}}  \\
\end{array}
\end{equation}
where $M(.)$ takes an image as input and output a person mask. $\bullet$ denotes a pixel-wise multiplication operator.

\textbf{\emph{Sub-domain Classification Loss.}}
To distinguish the sub-domain labels of a real/fake image , we use the cross entropy loss. The losses of real and fake images are computed by
\begin{equation}
\label{eq:dom}
\begin{array}{*{20}{l}}
   {L_{dom}^r({{\tilde D}_s},{{\tilde D}_t})} & { = \sum\limits_{i = 1}^M {\sum\limits_{j = 1}^N ( ({L_{ce}}({{\tilde D}_s}({x_{{s_i}}}),l_s^i))} }  \\
   {} & { + {L_{ce}}({{\tilde D}_t}({x_{{t_j}}}),l_t^j))}  \\
   {L_{dom}^f(G,\bar G,{{\tilde D}_s},{{\tilde D}_t})} & { = \sum\limits_{i = 1}^M {\sum\limits_{j = 1}^N ( ({L_{ce}}({{\tilde D}_s}(G({x_{{s_i}}},l_s^i,l_t^j)),l_t^j))} }  \\
   {} & { + {L_{ce}}({{\tilde D}_t}(G({x_{{t_j}}},l_t^j,l_s^i)),l_s^i))}  \\
\end{array}
\end{equation}
where the sub-domain classifiers ${\tilde D}_s$ and ${\tilde D}_t$ share the parameters with $D_s$ and $D_t$, respectively. $L_{ce}(.,.)$ denotes the cross entropy loss.

\textbf{\emph{Full objective.}} Considering all the loss described above, we decompose this two-player minimax optimization problem into two parts
\begin{equation}
\label{eq:solve}
\begin{array}{*{20}{l}}
   {{L_D}} & { =  - {L_{adv}}(G,{D_t}) - {L_{adv}}(\bar G,{D_s}) + {\lambda _1}L_{dom}^r({{\tilde D}_s},{{\tilde D}_t})}  \\
   {{L_G}} & { = {L_{adv}}(G,{D_t}) + {L_{adv}}(\bar G,{D_s})}  \\
   {} & { + {\lambda _1}L_{dom}^f(G,\bar G,{{\tilde D}_s},{{\tilde D}_t})}  \\
   {} & { + {\lambda _2}{L_{mask}}(G,\bar G) + {\lambda _3}{L_{rec}}(G,\bar G) }  \\
\end{array}
\end{equation}
where $\lambda _1$, $\lambda _2$ and $\lambda _3$ control the relative importance of the four objectives. We alternately optimize $L_D$ and $L_G$. When optimizing $L_D$, we fix $G$, $\bar G$ to optimize $D_s$, $D_t$, ${\tilde D}_s$, ${\tilde D}_t$, and vice versa.

\subsection{Feature Learning.}
After the transferring from source sub-domains to target sub-domains, we obtain lots of transferred fake images that follow the distributions of target sub-domains. With these transferred fake images and their corresponding raw identity labels, we can perform feature learning by using supervised person ReID methods. Note that we do not use any label in target domain. In this paper, we do not focus on how to learn a robust feature representation. Therefore, we simply use a conventional classification network (ResNet-50) as a base model.
\section{Implementation.}
\subsection{Network Architecture.}
Following \cite{CycleGAN2017}, M2M-GAN has two generators and two discriminators ( discriminators and classifiers share parameters). The generative networks consist of two stride-2 convolution layers, six residual blocks and two stride-1/2 fractionally-strided convolution layers. We also use instance normalization \cite{Instance2016} for generators. For discriminators, we use PatchGANs \cite{Isola_2017_CVPR}.

As for feature learning networks, we use a base ResNet-50, followed by a 512-dim FC layer, a Batch Normalization layer, a ReLU layer, a dropout layer, a $n$-dim FC layer ($n$ denotes the number of person identities) and a Cross Entropy layer. During testing, we extract the features from the Average Pooling layer.

\subsection{Training.}
When training the M2M-GAN networks, we use Adam with ${\beta _1}=0.5$ and  ${\beta _2}=0.999$. Following StarGAN \cite{Choi_2018_CVPR}, we perform one step of optimizing generators after five steps of optimizing discriminators. We train M2M-GAN with a learning rate of 0.0001 for the first 100,000 iterations and linearly decay the learning rate to 0 over the next 100,000 iterations. The batch size is set to 16. After training M2M-GAN, we generate $N$ fake images that follow $N$ target sub-domains respectively for each source image. The fake images are further used for feature learning. We set $\lambda_1=1$, $\lambda_2=100$, and $\lambda_3=10$.

When training the feature learning networks, the pre-trained ResNet-50 model is used and further fine-tuned on the translated target domain. We use SGD with mini-batch size of 256. The learning rate is initialized to 0.1 for FC layers and 0.01 for other layers. After 30 epochs, the learning rate is divided by 10. We resize images to $256\times 128$. We train these networks for 40 epochs. Note that only the training set of the source domain and unlabeled training set of the target domain are used for both M2M-GAN training and feature learning.

\section{Experiments}
\label{sec:exp}
In this section, we evaluate our M2M-GAN method on three large-scale person ReID benchmark datasets, i.e., Market-1501, DukeMTMC-reID and MSMT17, and present ablation studies to reveal the importance of each main component/factor of our method. We then reveal the benefits of the M2M-GAN model compared with state-of-the-art methods. We use CMC and mAP for evaluation.

\textbf{\emph{Datasets.} }
The Market-1501 dataset with six cameras is collected in Tsinghua University. Overlap exists among different cameras. Overall, this dataset contains 32,668 annotated bounding boxes of 1,501 identities. Among them, 12,936 images from 751 identities are used for training, and 19,732 images from 750 identities plus distractors are used for gallery. As for query, 3,368 hand-drawn bounding boxes from 750 identities are adopted. Each annotated identity is present in at least two cameras.

DukeMTMC-reID has 8 cameras. There are 1,404 identities appearing in more than two cameras and 408 identities (distractor ID) who appear in only one camera. Specially, 702 IDs are selected as the training set and the remaining 702 IDs are used as the testing set. In the testing set, one query image is picked for each ID in each camera and the remaining images are put in the gallery. In this way, there are 16,522 training images of 702 identities, 2,228 query images of the other 702 identities and 17,661 gallery images (702 ID + 408 distractor ID).

MSMT17 is the largest person re-identification dataset. It contains 15 cameras, i.e., 12 outdoor cameras and 3 indoor cameras. Four days with different weather conditions in a month are selected for video collection. For each day, 3 hours of videos taken in the morning, noon, and afternoon, respectively, are selected for pedestrian detection and annotation. The final raw video set contains 180 hours of videos, 12 outdoor cameras, 3 indoor cameras, and 12 time slots. Faster RCNN is utilized for pedestrian bounding box detection. Finally, 126,441 bounding boxes of 4,101 identities are annotated.

%

\begin{table*}[]
\begin{center}
\begin{tabular}{|l|c|c|c|c|c|c|c|c|c|c|c|c|}
\hline
\multirow{2}{*}{Methods} & \multicolumn{2}{c|}{D-\textgreater{}MA} & \multicolumn{2}{c|}{MA-\textgreater{}D} & \multicolumn{2}{c|}{D-\textgreater{}MS} & \multicolumn{2}{c|}{MS-\textgreater{}D} & \multicolumn{2}{c|}{MA-\textgreater{}MS} & \multicolumn{2}{c|}{MS-\textgreater{}MA} \\ \cline{2-13}
                         & R1                 & mAP                & R1                 & mAP                & R1                 & mAP                & R1                 & mAP                & R1                  & mAP                & R1                  & mAP                \\ \hline
supervised               & 85.4               & 66.9               & 76.5               & 57.1               & 61.3               & 30.8               & 76.5               & 57.1               & 61.3                & 30.8               & 85.4                & 66.9               \\ \hline
Pre-training             & 50.4               & 23.6               & 38.1               & 21.4               & 20.2               & 6.7                & 53.5               & 32.5               & 14.2                & 4.5                & 51.5                & 25.5               \\ \hline
one-to-one               & 47.4               & 21.5               & 43.1               & 24.1               & 24.7               & 7.8                & 51.1               & 30.0               & 22.7                & 7.6                & 46.1                & 21.1               \\ \hline
many-to-many (Ours)      & 57.5               & 26.8               & 49.6               & 26.1               & 35.3               & 11.2               & 56.6               & 33.3               & 30.9                & 10.1               & 53.4                & 25.2               \\ \hline
\end{tabular}

\end{center}
\caption{Effectiveness of M2M-GAN. There are six transferring combinations between three datasets. ``D", ``MA" and ``MS" denotes DukeMTMC-reID, Market-1501 and MSMT17, respectively.}\label{tab:one2one}
\end{table*}

\subsection{Evaluation and Model Analysis}
To evaluate the effectiveness of M2M-GAN, we conduct different kinds of ablation studies according to the CDTL setting.

\textbf{\emph{Effectiveness of M2M-GAN.}}
 To evaluate the effectiveness of M2M-GAN, one of three datasets is used as the target domain and the other two are used as source domains, respectively. Therefore, there are six transferring combinations.  We design several experimental comparisons to validate the points discussed in our paper. Note that we use the same feature learning networks and the same training scheme in all experiments.

We first validate the point that although cross-domain transfer learning is important in person ReID, it undergoes a serious degradation compared with supervised person ReID. We conduct two experiments according to these two methods, i.e., supervised person ReID (denoted as ``Supervised") and directly transferring (denoted as ``Directly transfer"). ``Supervised" is trained and tested on a labeled target domain. ``Directly transfer" is trained on a labeled source domain and tested on a target domain. As shown in Table \ref{tab:one2one}, we can see that the rank-1 accuracy of ``Directly transfer" is dropped by $\sim$30\% compared with ``Supervised", e.g., 35\%, 38.4\%, 41.1\%, 23.0\%, 47.1\%, and 33.9\% for six transferring processes, respectively. The reason is that there is a large distribution gap between two datasets collected in different scenarios.

We then validate the point that one-to-one CDTL method (denoted as ``One-to-one") is inappropriate for the CDTL of person ReID. Compared with ``Directly transfer", ``One-to-one" obtains rank-1 accuracy of -3.0\%, +5.0\%, 4.5\%, -2.4\%, 8.5\%, and -5.4\% improvement for six transferring processes, respectively. For three transferring processes, ``One-to-one" is better than ``Directly transfer" while for the other three transferring processes, ``Directly transfer" is even better than ``One-to-one". The reason may be attributed to the fact that the ignorance of sub-domains of the source/target domain confuses GANs, i.e., GANs do not know which sub-domain should be generated.


Most importantly, We further validate the point that our many-to-many CDTL method (denoted as ``Many-to-many") is better than both ``Directly transfer" and ``One-to-one" methods. Compared with ``Directly transfer", our ``Many-to-many" obtains rank-1 accuracy of 7.1\%, 11.5\%, 15.1\%, 3.1\%, 16.7\%, and 1.9\% for six transferring processes, respectively. These show that our ``Many-to-many" can reduce the domain gap between a source domain and a target domain and thus improve the performance of cross-domain person ReID. Compared with ``One-to-one", our ``Many-to-many" obtains 10.1\%, 6.5\%, 10.6\%, 5.5\%, 8.2\%, and 7.3\% for six transferring processes, respectively. These show that it is necessary to take camera-based sub-domains into consideration because different camera-based sub-domains in the same domain follow different distributions. When considering the source/target sub-domain labels into CDTL, GANs know which source sub-domain the image comes from and which target sub-domain the image will be translated and thus improve the performance of cross-domain person ReID.



\begin{figure}[!t]
\centering
\subfloat[Effect of $\lambda_1$.]{\includegraphics[width=1.6in,height=1.2in]{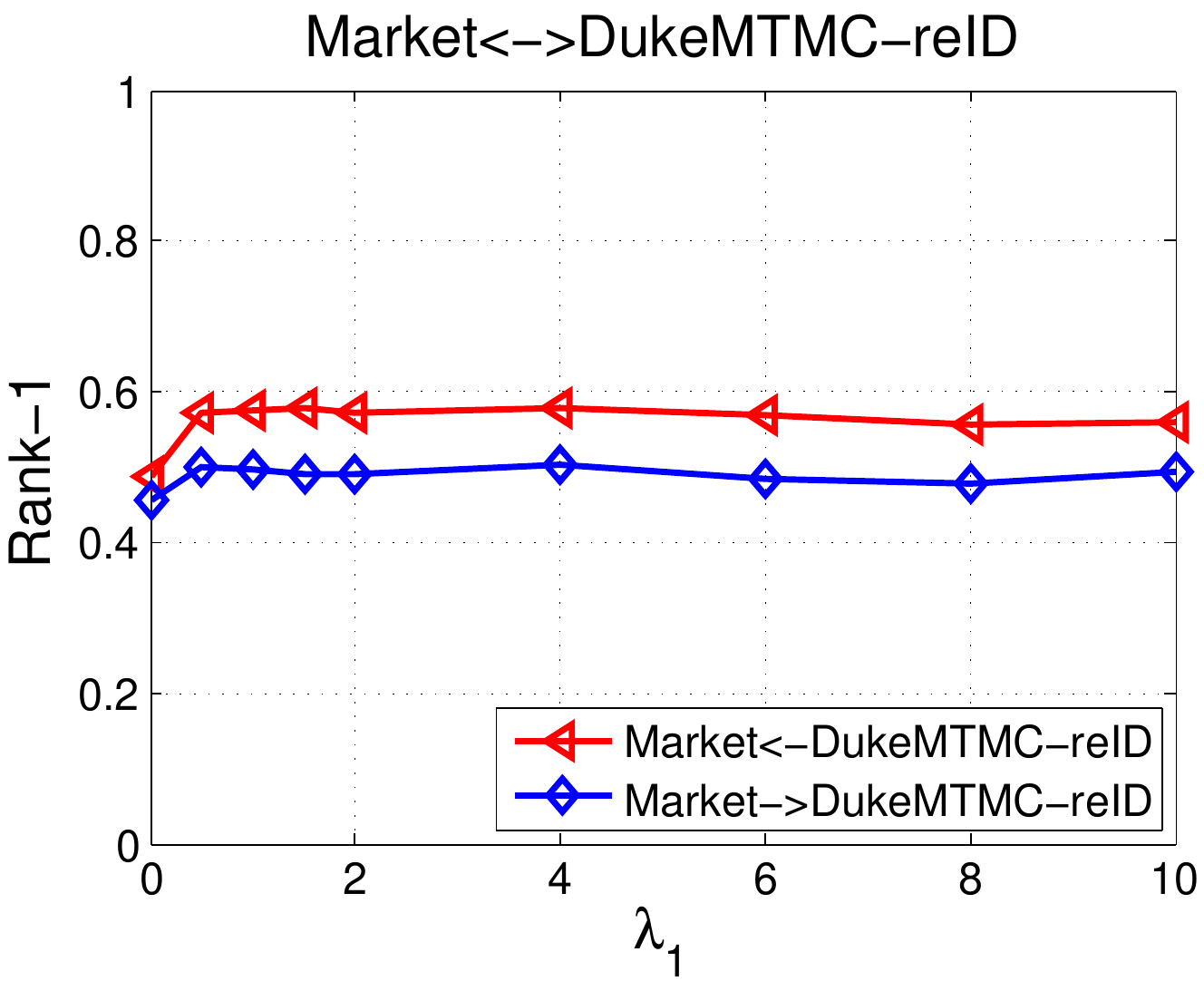}
\label{fig:lambda1}}
\hfil
\subfloat[Effect of $\lambda_2$.]{\includegraphics[width=1.6in,height=1.2in]{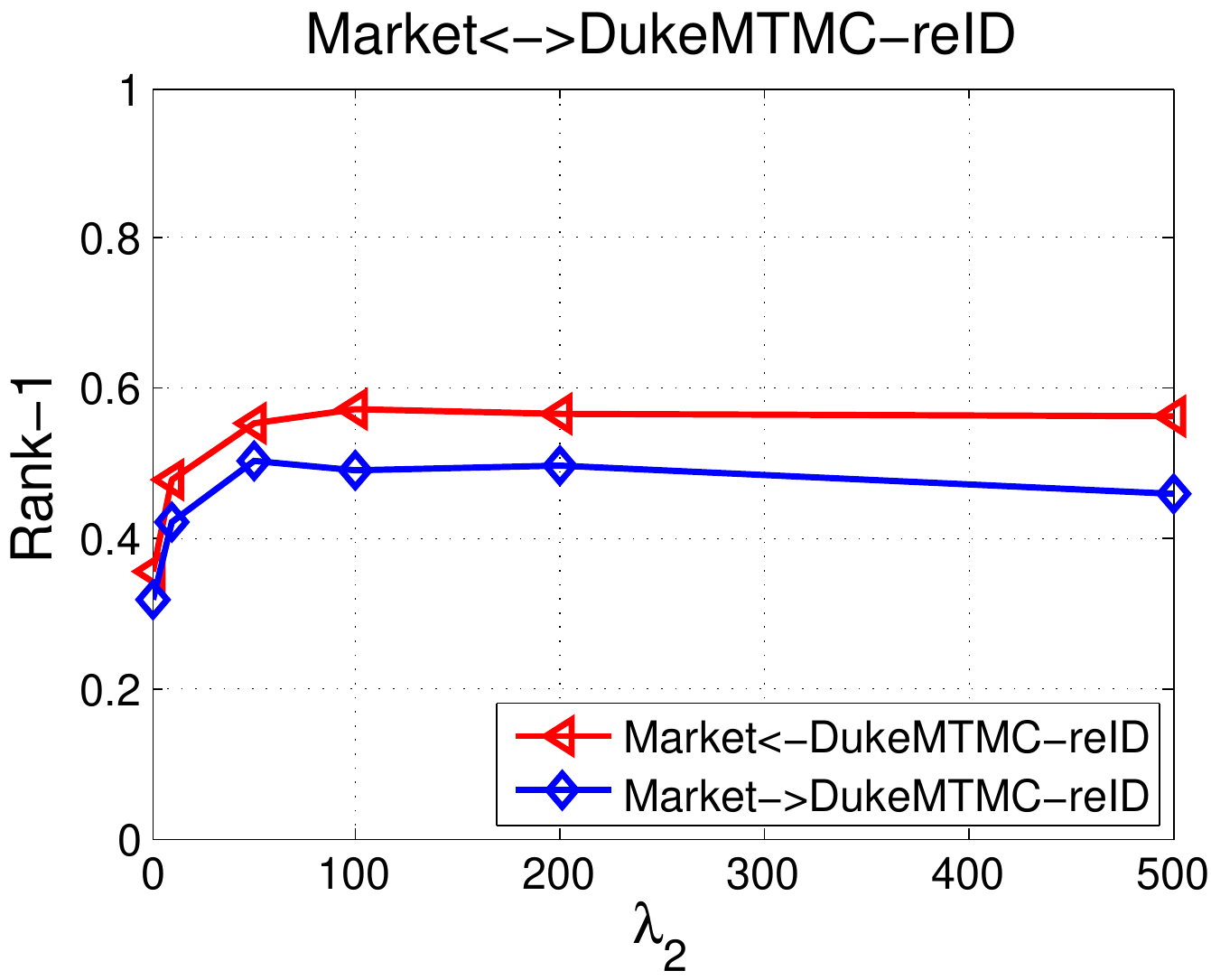}
\label{fig:lambda1}}
\hfil
\caption{Influence of parameters.}
\label{fig:parameters}
\end{figure}

\textbf{\emph{Influence of parameters.}} To investigate the impact of two important parameters in our M2M-GAN, i.e., $\lambda_1$ and $\lambda_2$, we conduct two sensitivity analysis experiments. As shown in Figure \ref{fig:parameters} (a) and (b), when $\lambda_1$ is in the range of 0.5$\sim$10.0 or $\lambda_2$ is in the range of 50$\sim$500, our model nearly keeps the best performance.

\begin{table*}[!t]
\begin{center}
\begin{tabular}{l|c|c|c|c|c|c|c|c|c|c|c|c}
\hline
\multirow{2}{*}{Methods} & \multicolumn{2}{c|}{D-\textgreater{}MA} & \multicolumn{2}{c|}{MA-\textgreater{}D} & \multicolumn{2}{c|}{D-\textgreater{}MS} & \multicolumn{2}{c|}{MS-\textgreater{}D} & \multicolumn{2}{c|}{MA-\textgreater{}MS} & \multicolumn{2}{c}{MS-\textgreater{}MA} \\ \cline{2-13}
                         & R1                 & mAP                & R1                 & mAP                & R1                 & mAP                & R1                 & mAP                & R1                 & mAP                 & R1                 & mAP                 \\ \hline
LOMO                     & 27.2               & 8.0                & 12.3               & 4.8                &-                    &-                    &12.3                &4.8                 &-                    &-                     &27.2                    &8.0                 \\
Bow                      & 35.8               & 14.8               & 17.1               & 8.3                &-                    &-                    &17.1                 &8.3                 &-                    &-                     &35.8           &  14.8          \\
EOLMA                 & 40.9               & -               & -               & -                &-                    &-                    &-                 &-                 &-                    &-                     &-           & -          \\ \hline

UMDL                     & 34.5               & 12.4               & 18.5               & 7.3                &-                    &-                    &                    &-                    &-                    &-                     &-                 &-              \\
PUL                      & 45.5               & 20.5               & 30.0               & 16.4               &-                    &-                    &-                 &-                    &-                    &-                     &-            &-                     \\
CAMEL                    & 54.5               & 26.3               & -                  & -                  &-                    &-                    &-                 &-                    &-                    &-                 &-                    &-                     \\ \hline

MMFA              & -               & -               & 45.3               & 24.7                &-                    &-                    &                    &-                    &-                    &-                     &-                 &-              \\
PTGAN                    & 38.6               & -                  & 27.4               & -                  &11.8        & 3.3                   &-                    &-                    &10.2        &  2.9                   &-                    &-                \\
SPGAN                    & 51.5               & 22.8               & 41.1               & 22.3               &-                    &-                    &-                    &-                    &-                    &-                     &-                    &-                     \\
SPGAN+LMP                    & 57.7               & 26.7               & 46.4       & 26.2                  &-                    &-                    &-                 &-                    &-                    &-                 &-                    &-                     \\
TJ-AIDL           & 58.2               & 26.5               & 44.3       & 23.0                  &-                    &-                    &-                 &-                    &-                    &-                 &-                    &-                     \\ \hline
M2M-GAN (Ours)           & 59.1               & 29.6               & 52.0       & 29.8                  &\textbf{36.8}                    &\textbf{11.9}                    &61.1                 &\textbf{37.5}                    &31.9        &\textbf{10.8}                 &57.9                    &28.8                     \\
\textbf{M2M-GAN+LMP (Ours)}                  &\textbf{63.1}                    &\textbf{30.9}                    &\textbf{54.4}                    &\textbf{31.6}                    &35.5                    &10.5                    &\textbf{61.7}                    &37.2                    &\textbf{32.2}                    &9.7                     &\textbf{60.8}                    & \textbf{29.7}                    \\ \hline

\end{tabular}
\end{center}
\caption{Comparisons to the State-of-the-Art. There are six transferring combinations between three datasets. ``D", ``MA" and ``MS" denotes DukeMTMC-reID, Market-1501 and MSMT17, respectively. The compared methods are categorized into three groups. Group 1: handcrafted feature methods. Group 2: unsupervised person ReID methods. Group 3: cross-domain transfer learning methods.}\label{tab:transfer3}
\end{table*}

\textbf{\emph{Effect of different feature learning schemes.}} To investigate the effect of different feature learning schemes, we conduct two experiments for evaluation. First, we use only fake target images and their corresponding source labels for training. Second, we use both fake target images and real source images with labels for training. The experimental results are shown in Table \ref{tab:feature_learning}. It is observed that although real source images do not follow the distribution of the target domain, it is still beneficial to the overall performance. The reason is that it is difficult to transfer all the information from a source domain to a target domain. Therefore, it is reasonable to re-use source images to compensate for the loss of information during transferring.
\begin{table}[]
\begin{center}
\begin{tabular}{|l|c|c|c|c|}
\hline
\multirow{2}{*}{Methods} & \multicolumn{2}{l|}{D-\textgreater{}MA} & \multicolumn{2}{l|}{MA-\textgreater{}D} \\ \cline{2-5}
                         & R1                 & mAP                & R1                 & mAP                \\ \hline
fake                     & 57.5               & 26.8               & 49.6               & 26.1               \\ \hline
fake+real                & \textbf{59.1}               & \textbf{29.6}               & \textbf{52.0}               & \textbf{29.8}               \\ \hline
\end{tabular}
\end{center}
\caption{Effect of different feature learning schemes.}\label{tab:feature_learning}
\end{table}

\textbf{\emph{Comparisons of computational cost.}} To analyze the computational cost of our M2M-GAN, we compare our M2M-GAN (denoted as ``M2M.") with a brute-force many-to-many method that contains $M\times N$ separate CycleGANs (denoted as ``Sper."), which is implicitly used in the recent work \cite{Wei_2018_CVPR}. As shown in Table \ref{tab:computation}, we can see that the model size of our ``M2M." nearly keeps the same with camera number increasing (the channels of first convolution kernel are affected) while the model size of ``Sper." is proportional to the camera number. When the camera number is large, ``Sper." is unavailable.

As for the training time, it is widely known that it costs much time to train a generative adversarial network because of an adversarial loss, e.g., half of a day. When training $M\times N$ separate CycleGANs on $M\times N$ pairs of source-to-target sub-domains, we estimate that our ``M2M." is much faster than ``Sper." empirically, even though each CycleGAN is trained on smaller sub-domains.


\begin{table}[]
\begin{center}
\begin{tabular}{|l|c|c|c|}
\hline
Meth.            & D\textless{}-\textgreater{}MA & MA\textless{}-\textgreater{}MS & MS\textless{}-\textgreater{}D \\ \hline
Sper. & 52.60M*8*6                     & 52.60M*6*15                     & 52.60M*15*8                    \\ \hline
M2M.            & \textbf{106.47M}                   & \textbf{106.58M}                 & \textbf{106.60M}                    \\ \hline
\end{tabular}
\end{center}
\caption{Comparisons of computational cost.}\label{tab:computation}
\end{table}

\subsection{Comparisons to the State-of-the-Art}
In this section, we compare our M2M-GAN with state-of-the-art methods by selecting one dataset as the target domain and the other two as the source domains, respectively. Therefore, we obtain six pairs of transferring combinations. The results are shown in Table \ref{tab:transfer3}.

We compare our approach with eleven state-of-the-art methods, which can be grouped into three categories. The first group includes three representative unsupervised person ReID methods using handcrafted features, i.e., LOMO \cite{liao2015person}, Bow \cite{zheng2015scalable} and EOLMA \cite{Zhou_2017_ICCV}. The experimental results clearly demonstrate the effectiveness of our M2M-GAN against the conventional handcrafted features, e.g., leading to 35.4\% and 42.1\% improvement on Market-1501 and DukeMTMC-reID respectively, compared with LOMO. The reason can be contributed to the fact that handcrafted features are inferior to the deep features.

The second group includes three unsupervised person ReID methods, i.e., UMDL \cite{Peng_2016_CVPR}, PUL \cite{Hehe17}, CAMEL \cite{Yu_2017_ICCV}. Compared with handcrafted features, the performance of these methods are improved to some extent. However, these methods do not consider the relationship between a labeled dataset and an unlabeled dataset and are therefore inferior to cross-domain methods.

The third group includes five representative CDTL methods, i.e., PTGAN \cite{Wei_2018_CVPR}, SPGAN \cite{Deng_2018_CVPR}, CAMEL \cite{Yu_2017_ICCV}, SPGAN+LMP \cite{Deng_2018_CVPR}, and TJ-AIDL \cite{Wang_2018_CVPR}. It is observed that our M2M-GAN outperforms all the competing methods. The main reason can be contributed the fact that our M2M-GAN takes camera-based source/target sub-domains into consideration and optimized in a unified process.

\section{Conclusion}
\label{sec:con}
In this paper, we propose a novel Many-to-Many Generative Adversarial Transfer Learning method (M2M-GAN) takes multiple source sub-domains and multiple target sub-domains into consideration and performs each pair-wise sub-domain transferring from the source domain to the target domain in a unified optimization process. As a result, our M2M-GAN solve the many-to-many CDTL with less training time, fewer parameters and a better performance. Experimental results on three large-scale benchmark datasets show the effectiveness of our M2M-GAN. We intend to extend this work in two directions. First, we intend to generalize the M2M-GAN to a generalized CDTL problem with any kind of adversarial topology. Second, the sub-domains of M2M-GAN are not limited to the camera-based sub-domains and therefore M2M-GAN can be easily generalized to any many-to-many CDTL problem.

\bibliography{egbib}
\bibliographystyle{aaai}

\end{document}